\documentclass{article}
\usepackage{spconf,amsmath,graphicx}
\usepackage{makecell}
\usepackage{booktabs}

\title{Abnormal Behavior Detection Based on Target Analysis}
\name{Luchuan Song, Bin Liu, Huihui Zhu, Weihai Li and Nenghai Yu}
\address{School of Information Science and Technology, University of Science and Technology of China, \\ Key Laboratory of Electromagnetic Space Information, Chinese Academy of Science}
\begin{document}
%
\maketitle
\vspace{-0.2cm}
\begin{abstract}
\vspace{-0.2cm}
Abnormal behavior detection in surveillance video is a pivotal part of the intelligent city. Most existing methods only consider how to detect anomalies, with less considering to explain the reason of the anomalies. We investigate an orthogonal perspective based on the reason of these abnormal behaviors. To this end, we propose a multivariate fusion method that analyzes each target through three branches: object, action and motion. The object branch focuses on the appearance information, the motion branch focuses on the distribution of the motion features, and the action branch focuses on the action category of the target. The information that these branches focus on is different, and they can complement each other and jointly detect abnormal behavior. The final abnormal score can then be obtained by combining the abnormal scores of the three branches.
\end{abstract}
\vspace{-0.1cm}
\vspace{-0.2cm}
\vspace{-0.2cm}
\section{Introduction}
\vspace{-0.2cm}
\label{sec:intro}

Abnormal behavior detection in surveillance video is a challenging task in computer vision. In practical applications, the definition of abnormality in the video is varied. For example, it may be an abnormal event for a pedestrian to ride a bicycle in the square, but cycling on a non-motor vehicle lane is usually regarded as a normal behavior. Therefore, it is difficult to define abnormality. Generally speaking, a video event is usually considered as an anomaly if it is not very likely to occur in the video \cite{cong2011sparse}. Unlike other tasks, there are usually only normal samples given by the training set and the anomalies in the environment are defined by these normal samples.\par
With the development of intelligent monitoring, we not only need to accurately detect the abnormal area in the surveillance video, but also need to analyse the reasons of anomalies. 
\begin{figure}[htp]
	\centering
	\includegraphics[width=0.5\textwidth]{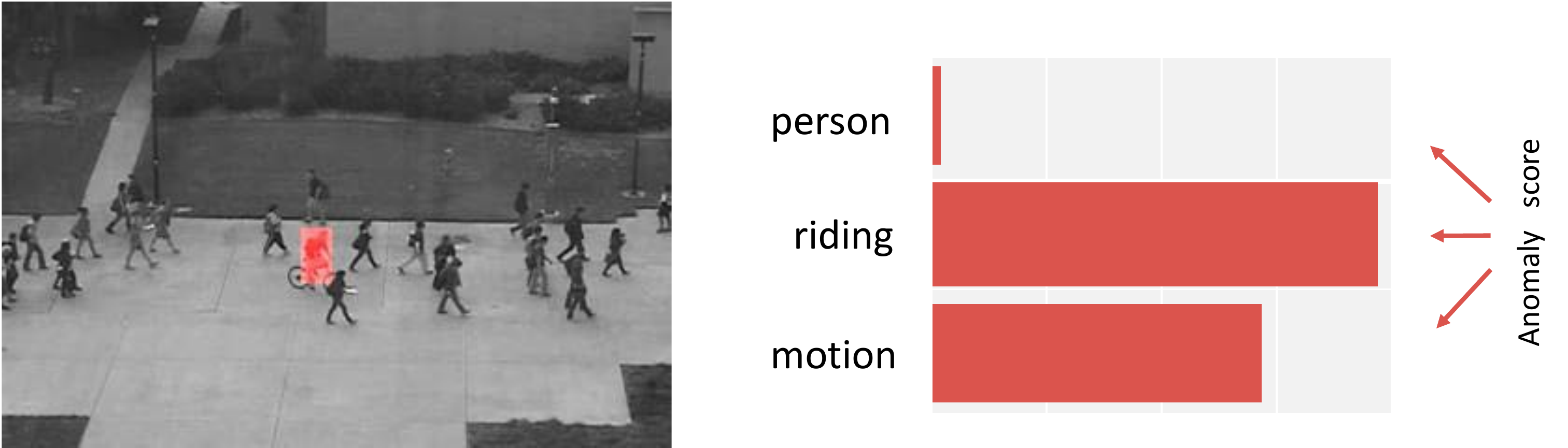}
	\vspace{-0.4cm}
	\caption{Examples of detect anomalies and explain anomalies through three branches: object, action and motion.}
	\vspace{-0.4cm}
	\label{explain}
\end{figure}
If we can give the reasons for the anomalies when we detect it, it would help the observers quickly judge whether they are false alarms or not. There are various factors that may be considered as the abnormal behaviors, such as holding a gun, running in a panic and so on. Therefore, if we want to explain the abnormal behavior, we need to analyze the target from multiple perspectives. Through the fusion result of each perspective, we can jointly detect anomalies and explain them.\par
Here we propose a multivariate fusion method to detect and explain anomalies. We analyze each target through three perspectives: object, action and motion. The object branch focuses on the appearance information, the motion branch focuses on the distribution of motion features, and the action branch focuses on the action category of the target. Both motion and action perspectives use video information.
The focus of the three perspectives is different, but they could complement each other. When anomalies are detected, the reason for determining the anomalies can be explained. As shown in Fig.\ref{explain}, the red area on the left is a person riding with abnormal motion, which is regarded as an abnormal event in the UCSD Ped2 dataset. On the right is the detection result of the target, which is explained as 'person', 'riding', 'abnormal motion'.\par
However, how to detect the action category of each target in a video is a critical task. Most of the existing action recognition algorithms only classify videos instead of targets. In a video, a few people continuously perform an action, and then the entire video is classified into some action category, which is called single-target and single-action recognition. In the anomaly detection field, each frame of video has multiple targets, and every target performs different actions. Therefore, these algorithms cannot solve the multi-target and multi-action recognition problem.\par
We propose an action recognition module to solve this problem. First, each frame is sent to the object detection network to obtain the category and coordinates of every target. Then, each target is tracked by visual tracking algorithm to obtain the target of the subsequent frame. Combined with object detection and visual tracking, we can get a video that contains only one target. 
\begin{figure}[htp]
	\centering
	\includegraphics[width=0.5\textwidth]{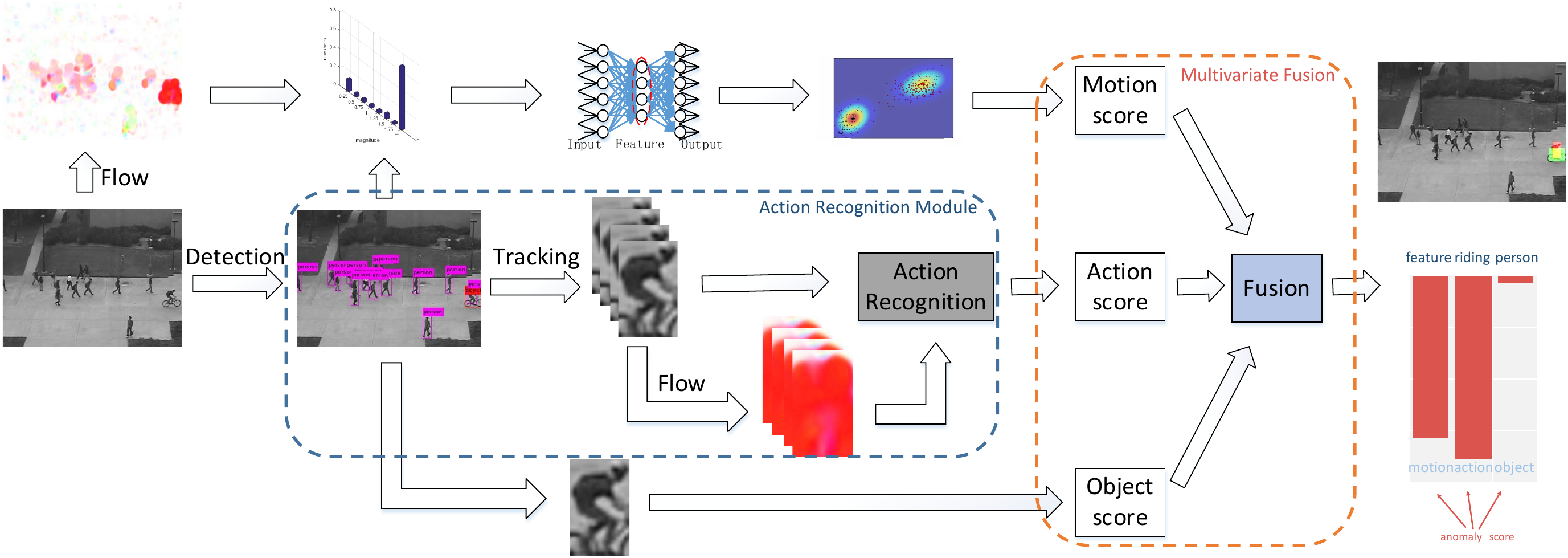}
	\vspace{-0.4cm}
	\caption{Framework of the proposed method.}
	\vspace{-0.5cm}
	\label{fig1}
\end{figure}
Taking these videos as input of the action recognition network to obtain the category of the action and the confidence score. The main contributions of our work are as follows:\par
(1) We propose a multivariate fusion method to detect and explain anomalies through three branches: object, action and motion. The performance of the fused method is not only higher than the performance of each branch, but also outperforms the state-of-the-art methods.\par
(2) We propose an action recognition module to solve the multi-target and multi-action recognition problem in the surveillance video. As far as we know, this is the first time that the action recognition using inter-frame information is utilized for anomaly detection.\par

\vspace{-0.35cm}
\section{Related Works}
\label{sec:format}
\vspace{-0.2cm}

In early studies, the traditional methods based on model-driven played a dominant role in anomaly detection. These methods usually extract hand-crafted features, which are generally divided into two types: motion and appearance. Motion-based features are often classified into three categories. The first is based on optical flow, such as Histograms of Oriented Optical Flow (HOF) \cite{xu2014video}, Multi-scale Histogram of Optical Flow (MHOF) \cite{cong2011sparse} and Histogram of Magnitude Optical Flow (HMOF). The second is based on trajectories \cite{jiang2011anomalous}. They focus on how to learn the normal trajectories of targets in videos. The last category is based on energy, which considers the crowd density and energy distribution, such as social force model \cite{mehran2009abnormal}, pedestrian loss model \cite{scovanner2009learning}. Appearance-based features generally include RGB information, such as Histogram of Oriented Gradient (HOG) \cite{dalal2005histograms} and Spatio-temporal Gradients \cite{kratz2009anomaly}.\par
In recent years, anomaly detection algorithms represented by deep learning are also developed and achieving good results. These methods are data driven and usually do not extract hand-crafted features. Instead, it uses neural networks to extract high-level features from video sequences \cite{feng2016deep}. For example, Sabokrou et al. \cite{sabokrou2018deep} used Fully Convolutional Neural Network (FCNs) to extract deep features to distinguish anomalies. However, these methods did not utilize the video information. As we all know, action is a continuous behavior, which means that inter-frame information is more important. Although MT-FRCN analyzed each target from three perspectives, the redundancy among them is high by only using the single-frame information. 
\begin{figure}[htp]
	\centering
	\includegraphics[width=0.5\textwidth]{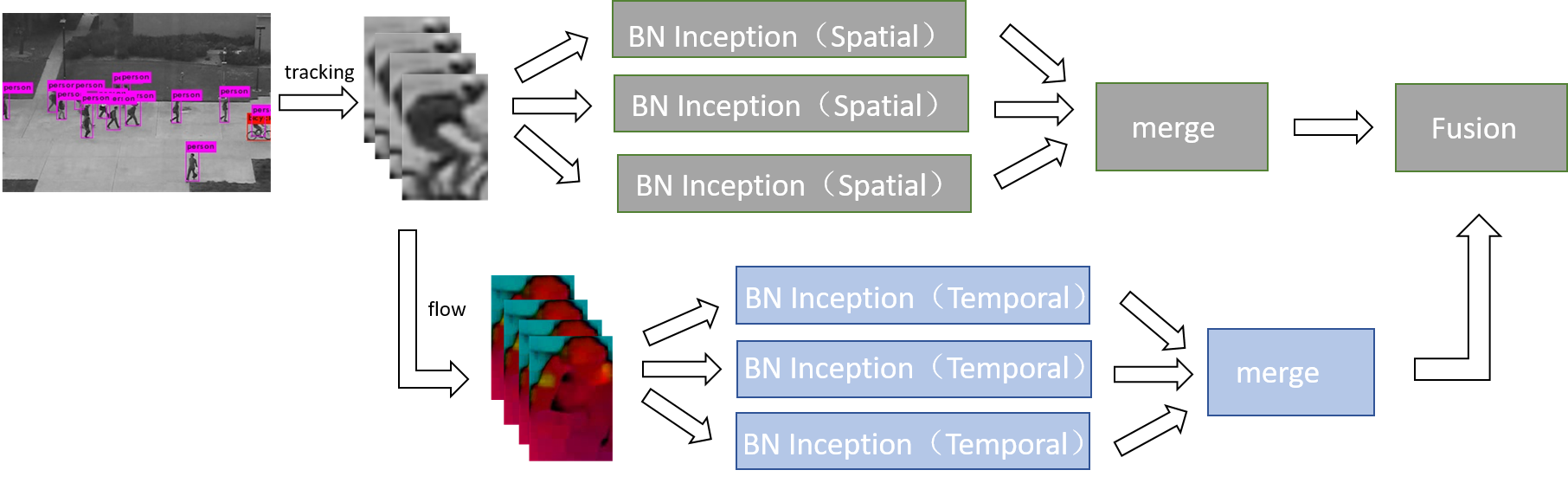}
	\vspace{-0.3cm}
	\caption{The process of action recognition.}
	\vspace{-0.3cm}
	\label{action}
\end{figure}
Compared to each perspective, the performance of the fusion was degraded, which indicated that this method has some limitations.

\vspace{-0.5cm}
\section{Methodology}
\label{sec:pagestyle}
\vspace{-0.2cm}

The proposed method is shown in Fig.\ref{fig1}. Firstly, the frame is sent to the object detection network to obtain the label and the coordinates of each target. Then, we analyze each target through three branches: object, action and motion. Finally, we can get the final abnormal score by combining the three branches. When the anomalies are detected, the reasons for determining the anomalies can be explained.\par

\vspace{-0.4cm}
\subsection{Object Branch}
\vspace{-0.2cm}

The first step is to detect every targets in the video. We use the well-known Yolo v3 \cite{redmon2018yolov3} algorithm for object detection. Each frame of video is inputted into the Yolo v3 to obtain the label, confidence score, and coordinates of the targets. We only keep the label with highest confidence score.\par
In the anomaly detection, the training set usually only contains the normal samples, thus we don't know which labels correspond to the anomalies. According to the definition of the anomalies, an event is usually considered as an anomaly if it is not very likely to occur in the video. Therefore, if the label of the object appears in the training set, high object confidence means the low probability of anomalies. Otherwise, high object confidence means that the target is more likely to be an anomaly. Considering that the network may obtain some false detections in the object label of the training set, we only retain the object label with high confidence.\par
We take the training set of the anomaly detection dataset as the input of the Yolo v3, and get the label and confidence of each target. We choose the label whose confidence is higher than a threshold \emph{$\alpha$} to store in a list denoted as $label\_list$. The abnormal score $Sco_{obj}$ of each target is defined as:
\begin{equation}
Sco_{obj}=\left\{ \begin{array}{l}\!-\!Conf_{obj}\hspace{0.7cm}if \hspace{0.1cm}Obj_{label}\hspace{0.1cm}in\hspace{0.1cm}label\_list\\
Conf_{obj}\hspace{0.8cm}otherwise
\end{array} \right.\
\end{equation}
where $Obj_{label}$ and $Conf_{object}$ are the label and the confidence of the target respectively. The $Sco_{obj}$ of all targets in the test set are stored into a list denoted as $Sco_{obj}\_list$,
and then we normalize each abnormal score to [0,1].
\begin{equation}
Sco_{obj} = \frac{{Sco_{obj} - \min (Sco_{obj}\_list)}}{{\max (Sco_{obj}\_list) - \min (Sco_{obj}\_list)}}
\end{equation}

\vspace{-0.3cm}
\subsection{Action Branch}

The difficulty for action recognition in the anomaly detection field is how to solve the problems of multi-target and multi-action recognition in a surveillance video. We propose a action recognition module to solve this problem. Based on the object detection, we use the well-known KCF \cite{henriques2015high} to track the target. Each target is tracked in \emph{$K$} frames, so we can get the the coordinates of the  continuous \emph{$K$}-frame of the target. After clipping, we get a short RGB video only containing the target. At the same time, the optical flow is extracted by using the coordinates of each target through previous frame and current frame of the video and obtain the optical flow video. Taking RGB video and optical flow video as the input of the action recognition network to obtain the action category and confidence of the target, thus we can solve the multi-target and multi-action detection problem in anomaly detection.\par
As for the action recognition network, we borrow the idea from the TSN algorithm \cite{wang2016temporal}. The TSN algorithm combines a sparse temporal sampling strategy and video-level supervision to enable efficient and effective learning using the whole action video. The video used in the TSN algorithm usually contains much redundant information. However, in the anomaly detection tasks, the video used for the action recognition is a short video obtained from the target tracking, which contains less redundant information. Therefore, we remove the sparse temporal sampling strategy.\par
The process of the action branch is shown in Fig.\ref{action}. We use a two-stream ConvNet architecture which incorporates spatial and temporal networks for action recognition. The ConvNet used in our work is BNInception network \cite{ioffe2015batch}. We send each RGB image and optical flow image of the short video into the spatial network and the temporal network respectively and the category scores of different frames are averaged. Finally the category scores of the RGB and optical flow branches are fused by a weight. The weight and the training process is similar to those in the TSN algorithm.\par
Similar to the object branch, if the action category of the target has appeared in the training set, high action confidence means low probability of anomalies.
We take the video frame sequence of each target tracked in the training set as the input of the action recognition network, and get the action category and confidence. After that, we choose the action category whose confidence is higher than \emph{$\beta$} to store in a list denoted as $action\_list$. The action abnormal score $Sco_{act}$ of each target is defined as:
\begin{equation}
Sco_{act}=\left\{ \begin{array}{l}-Conf_{act}\hspace{0.4cm}if \hspace{0.1cm}Act_{label}\hspace{0.1cm} in\hspace{0.1cm} action\_list\\
Conf_{act}\hspace{0.7cm}otherwise
\end{array} \right.\
\end{equation}
where $Act_{label}$ and  $Conf_{act}$ are the action category and the confidence of the target respectively.
The action abnormal score $Sco_{act}$ of all targets in the test set are stored into a list denoted as $Sco_{act}\_list$, and then we normalize each abnormal score to [0,1].
\begin{equation}
Sco_{act} = \frac{{Sco_{act} - \min (Sco_{act}\_list)}}{{\max (Sco_{act}\_list) - \min (Sco_{act}\_list)}}
\end{equation}

\vspace{-0.3cm}
\subsection{Motion Branch}
\vspace{-0.1cm}

We use Histogram of Magnitude Optical Flow (HMOF) features as the motion features. We extract the HMOF features of each target in the video and then send them to the auto-encoder network for reconstruction in order to further expand the difference between normal and abnormal features. All features of the training set are used to train the Gaussian Mixture Model (GMM) Classifiers, then we use the trained classifier to test the features of the testing set. Each feature get a score after passing through the classifier. The score is denoted as $Sco_{mot}$. Finally, the motion abnormal score $Sco_{mot}$ of all targets in the test set are stored into a list denoted as $Sco_{mot}\_list$, and each abnormal score is normalized to [0, 1].
\begin{equation}
Sco_{mot} = \frac{{\max (Sco_{mot}\_list) - Sco_{mot} }}{{\max (Sco_{mot}\_list) - \min (Sco_{mot}\_list)}}
\end{equation}
The larger the $Sco_{mot}$, the less possibility the motion feature of the target match the distribution of normal motion features, so the target is more likely to be an anomaly.

\vspace{-0.3cm}
\subsection{Fusion}
\vspace{-0.2cm}

After getting the abnormal scores of the three branches, we fuse these abnormal scores to get the final abnormal score.
However, different branches play a different role in various scenes. For example, in the UMN dataset, the scene is more concerned with the panic running of the crowd, the weights of the motion and action branches should be higher than the object branch. Thus, the abnormal score of each branch should be weighted by a factor, and the final abnormal score of the target is determined by the three weighted abnormal scores.\par
In addition, the anomaly detection is different from other tasks. In practical applications, the monitoring system notifies the staff to handle the abnormality after detecting the abnormality. A few false detections only slightly burden the staff, but a few missing detections mean that the abnormal events cannot be handled in time, which may eventually lead to serious consequences.
Considering the practical significance of anomaly detection, we select the maximum value of the abnormal scores from the weighted branches as the final fusing score denoted as $Sco_{fuse}$:
\begin{equation}
\max ({w_{obj}}*Sco_{obj},{w_{act}}*Sco_{act},{w_{mot}}*Sco_{mot})
\end{equation}
where the $w_{obj}$, $w_{act}$, $w_{mot}$ are the weight of the object, action and motion branch respectively.
The $Sco_{fuse}$ of all targets in the test set are stored into a list denoted as $Sco_{fuse}\_list$,
and then we normalize the final abnormal score $Sco_{fuse}$ of each target to [0,1].
\begin{equation}
Sco_{fuse} = \frac{{Sco_{fuse} - \min (Sco_{fuse}\_list)}}{{\max (Sco_{fuse}\_list) - \min (Sco_{fuse}\_list)}}
\end{equation}
If the $Sco_{fus}$ is greater than the threshold \emph{$\partial$}, it is considered an anomaly.
\begin{equation}
object = \left\{ \begin{array}{l}
abnoraml\hspace{0.7cm}if\hspace{0.1cm}Sco_{fus} > \partial \\
normal\hspace{1cm}otherwise
\end{array} \right.
\end{equation}

\begin{figure}[htp]
	\centering
	\includegraphics[width=0.5\textwidth]{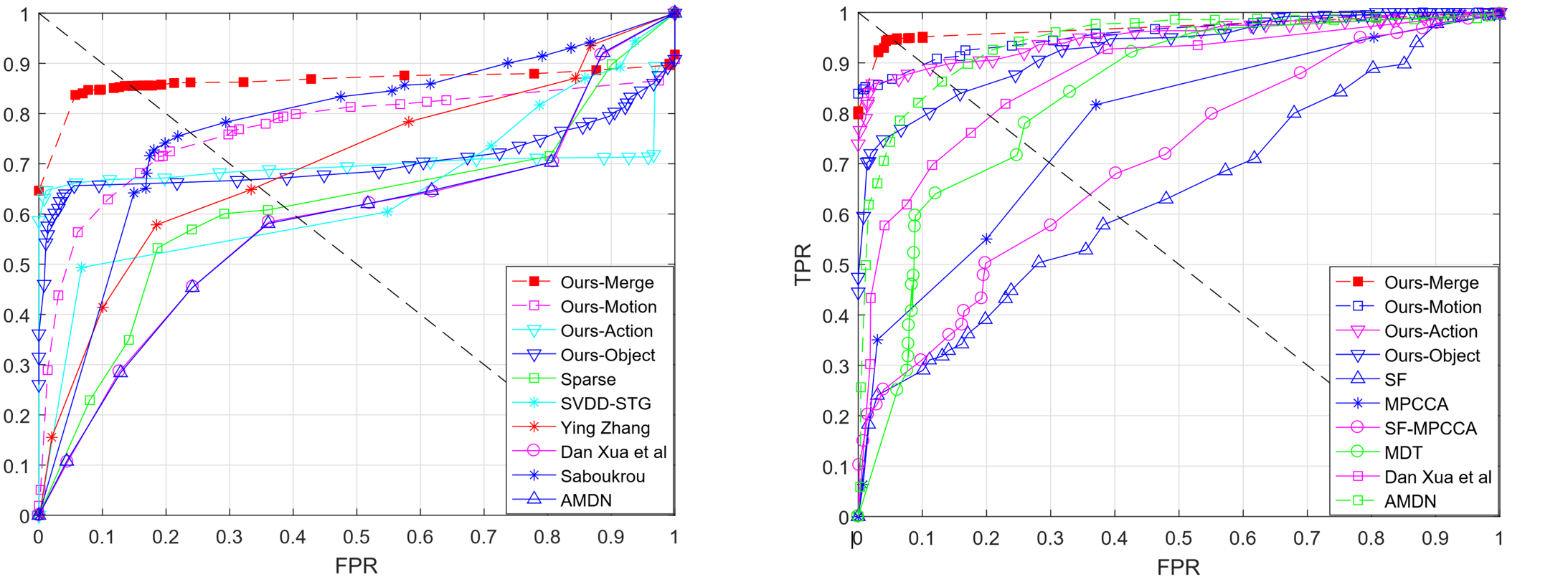}
	\vspace{-0.5cm}
	\caption{ROC comparison with state-of-the-art methods on UCSD Ped2 dataset. Left: Frame-level. Right: Pixel-level}
	\vspace{-0.5cm}
	\label{figroc}
\end{figure}
\vspace{-0.5cm}
\section{Experiments}
\label{sec:intro}
\vspace{-0.3cm}

We provide the representative experimental results on two datasets: the UMN dataset and the UCSD Ped2 dataset. In the action branch, we select about 8,000 videos with 16 common action categories from different commom datasets for training.\par
As for the evaluation criteria, we adopt the frame-level criteria for anomaly detection and the pixel-level criteria for anomaly localization \cite{sabokrou2015real}. For frame-level, if one pixel is detected as an anomaly, the whole frame is considered as an anomaly. For pixel-level, if more than 40\% of truly abnormal pixels are detected, it will be treated as true positive \cite{sabokrou2015real}. Furthermore, two criteria are used to evaluate the ROC curves: Area Under Curve (AUC) and Equal Error Rate (EER). Higher AUC and lower EER indicate a better performance.

\vspace{-0.4cm}
\subsection{UMN Dataset}
\vspace{-0.2cm}
Because the UMN dataset does not have pixel-level annotation information, we only evaluated it on the frame-level.\par
The hyperparameters are empirically set as follows: ${\rm{w}}_{object}$, ${\rm{w}}_{action}$, ${\rm{w}}_{object}$ are set to 1, 1.5, 1.5. \emph{$\alpha$} is 0.95, and \emph{$\beta$} is 0.99. The threshold \emph{$\vartheta $} of HMOF is 1.8, the interval \emph{$n$} is 8, and the number \emph{$K$} of the tracking frames is 5.\par
Table.\ref{table1} shows the performance comparison between the proposed algorithm and other state-of-the-art algorithms. The object branch's performance is very poor, since the anomalies are panic people, and the object branch is difficult to detect the anomaly. The motion branch is sensitive to the panic people and has better performance. The performance of the proposed method after fusion is not reduced, which indicates the effectiveness of the method.
Compared with other state-of-the-art algorithms, the proposed algorithm has the best performance in the UMN dataset.
\begin{table}[htp]
	\centering
	\begin{tabular}{lc|lc}  
		\toprule[2pt]
		Method                           &AUC                & Method   &AUC       \\ \hline  
		ZH et al.\cite{liu2014abnormal}       &99.3\%             &Leyva et al.\cite{leyva2017video}         &88.3\%          \\         
		SR\cite{cong2011sparse}    &97.5\%                 &Ours-object       &48.6\%           \\ 
		MIP\cite{du2013abnormal}             &94.4\%                &Ours-action    &84.8\%                     \\         
		Sabokrou et al.\cite{sabokrou2015real}    &99.6\%              &Ours-motion     &{\bf 99.8\%}           \\         
		DeepCascade\cite{sabokrou2017deep}         &99.6\%             &Ours-fusion     &{\bf 99.8\%}         \\         
		\bottomrule[2pt]
	\end{tabular}
	\vspace{-0.2cm}
	\caption{AUC comparison of our proposed method with the state-of-the-art methods in the UMN dataset.}
	\vspace{-0.5cm}
	\label{table1}
\end{table}
\vspace{-0.3cm}
\subsection{UCSD Dataset}

The UCSD Ped2 dataset is the standard benchmark for anomaly detection, and have different scenes.
The hyperparameters are empirically set as follows:
${\rm{w}}_{object}$, ${\rm{w}}_{action}$, ${\rm{w}}_{object}$ are set to 1, \emph{$\alpha$} is 0.95, and \emph{$\beta$} is 0.99. The threshold \emph{$\vartheta $} of HMOF is 2.4, the interval \emph{$n$} is 8, and the number \emph{$K$} of tracking frames is 5.\par
Fig.\ref{figroc} shows the ROC curves on the UCSD Ped2 dataset and the EER comparison with state-of-the-art method is shown in Table\ref{table2}. From Fig.\ref{figroc} and Table.\ref{table2}, we can see that:\par
1.  Object, motion and action branches's performance have their own limitations, but the performance of the fusion is better than the respective branches in frame-level and pixel-level.\par
2. The performance of our method is significantly improved compared with MT-FRCN. The combined performance EER decreased by 11.6\% at the frame-level, from 17.1\% to 5.5\%, and decreased by 4.9\% at the pixel-level, from 19.4\% to 14.5\%.\par
3. Compared with other state-of-the-art algorithms, EER of our method has reached the lowest at frame-level and pixel-level respectively.\par


\begin{table}
	\centering
	\linespread{0.2}
	\begin{tabular}{ccccccccc}
		\toprule[2pt]
		Method                                     &Pixel-level&Frame-level \\
		\hline
		MDT\cite{mahadevan2010anomaly}             &24\%&54\% \\
		Zhang et al.\cite{zhang2016combining}      &22\%&33\% \\
		Xua et al.\cite{xu2014video}               &20\%&42\% \\
		IBC\cite{boiman2007detecting}              &13\%&26\% \\
		Li et al.\cite{li2014anomaly}              &18.5\%&29.9\% \\
		Leyva et al.\cite{leyva2017video}          &19.2\%&36.6\% \\
		Sabokrou et al.\cite{sabokrou2015real}     &19\%&24\% \\
		Xiao et al.\cite{xiao2015learning}         &10\%&17\% \\
		Deep-anomaly\cite{sabokrou2018deep}        &11\%&15\% \\
		MT-FRCN\cite{hinami2017joint}              &17.1\%&19.4\% \\
		\emph{Ours-object}                         &16.0\%&33.2\% \\
		\emph{Ours-action}                         &11.1\%&31.5\% \\
		\emph{Ours-motion}                         &10.6\%&25.3\% \\
		\emph{Ours-fusion}                         &{\bf 5.5\%}&{\bf 14.5\%} \\
		\bottomrule[2pt]
	\end{tabular}
	\vspace{-0.2cm}
	\caption{EER comparison of our proposed method with state-of-the-art methods in the UCSD Ped2 dataset.
	}
	\vspace{-0.1cm}
	\label{table2}
\end{table}

\vspace{-0.5cm}
\section{Conclusion}
\label{sec:intro}
\vspace{-0.3cm}
This paper proposes a multivariate fusion method based on target analysis, which not only focuses on how to detect anomalies, but also try to explain the reason of anomalies. We analyze each target through three branches: object, action and motion. The information that these branches focus on is different, and they can jointly detect and explain anomalies. Furthermore, in the action branches, we propose an action recognition module using inter-frame information to solve the multi-target and multi-action recognition problems, which has not been utilized in anomaly detection field before.





\bibliographystyle{IEEEbib}
\bibliography{strings,refs}

\end{document}